\documentclass[preprint,12pt]{elsarticle}
\usepackage{longtable}

\usepackage{hhline}
\usepackage{graphicx}
\usepackage{mathrsfs}
\usepackage{times}
\usepackage{rotating}
\usepackage{subfigure}
\usepackage{latexsym}
\usepackage{amsmath}
\usepackage{algorithm}
\usepackage{algorithmic}
\usepackage{color}
\usepackage{syntonly}
\usepackage{amsfonts}

\usepackage{amssymb}

\journal{Pattern Recognition}

\begin{document}

\begin{frontmatter}



\title{Curvature-aware Manifold Learning}

 
 \author[mymainaddress,mysecondaryaddress]{Yangyang Li\corref{mycorrespondingauthor}}
 \cortext[mycorrespondingauthor]{Corresponding author}
 \ead{liyangyang12@mails.ucas.ac.cn}
\address[mymainaddress]{Academy of Mathematics and Systems Science Key Lab of MADIS\\
	Chinese Academy of Sciences, Beijing, 100190, China}
\address[mysecondaryaddress]{University of Chinese Academy of Sciences, Beijing, 100049, China}

\begin{abstract}
Traditional manifold learning algorithms assumed that the embedded manifold is globally or locally isometric to Euclidean space. Under this assumption, they divided manifold into a set of  overlapping local patches which are locally isometric to linear subsets of Euclidean space. By analyzing the global or local isometry assumptions it can be shown that the learnt manifold is a flat manifold with zero Riemannian curvature tensor. In general, manifolds may not satisfy these hypotheses. One major limitation of traditional manifold learning is that it does not consider the curvature information of manifold. 
In order to remove these limitations, we present our curvature-aware manifold learning algorithm called CAML. The purpose of our algorithm is to break the local isometry assumption and to reduce the dimension of the general manifold which is not isometric to Euclidean space. Thus, our method adds the curvature information to the process of manifold learning. The experiments have shown that our method CAML is more stable than other manifold learning algorithms by comparing the neighborhood preserving ratios.

\end{abstract}


\end{frontmatter}


\section{Introduction}
\label{}
In many machine learning tasks, one is often confronted with the redundant dimension of data points. There is a strong intuition that the data points may have an intrinsic lower dimensional representation. The concept of manifold was first applied in dimension reduction in \cite{3} and \cite{6}, called manifold learning (MAL). In this decade, manifold learning has become a significant component of machine learning, pattern recognition, image vision and so on. Traditional manifold learning algorithms aim to reduce the dimensionality of high dimensional data points, so that the lower dimensional representations could reflect the intrinsic geometrical and topological structure of the high dimensional sampled points. In general, the existing manifold learning algorithms are mainly divided into two classes: global and local \cite{21}. Global approaches aim to preserve the global geometric structure of the manifold during dimension reduction, such as IsoMap \cite{3}; Local approaches attempt to uncover the geometric structures of local patches, such as LLE \cite{6}, LEP \cite{1}, LPP \cite{4}, LTSA \cite{5}, Hessian Eigenmap \cite{2} et al. Isomap aims to preserve the geodesic distance between any two high dimensional data points, which can be viewed as a nonlinear extension to Multidimensional Scaling (MDS) \cite{11}. Locally preserved manifold learning algorithms aim to inherit and preserve the local geometric structure of embedded manifold. For instance, LLE aims to preserve the local linear structures of local patches and LEP aims to preserve the local similarities among data points during dimension reduction.

\begin{sidewaystable}
	\centering  
	\begin{tabular}{|l|c|c|c|}  
		\hline
		Authors &Year &Algorithm &Manifold Assumption  \\ \hline
		Tenenbaum et al. &2000 &IsoMap &Globally isometric to a convex subset of Euclidean space \\ \hline
		Roweis et al. &2000 &LLE &Locally linear  \\ \hline
		Belkin et al &2003 &LEP &Locally linear  \\ \hline
		Donoho et al. &2003 &HLLE &Locally isometric to an open, connected subset of Euclidean space \\ \hline
		Zhang et al. &2004 &LTSA &Locally linear \\ \hline
		He et al. &2005 &LPP &Linear form of Laplacian eigenmaps\\ \hline
		Dollar et al. &2007 &LSML & Not locally isometric to Euclidean space \\ \hline
		Binbin Lin et al. &2013 &PFE & Local isometric to subspace of Euclidean space\\ \hline
	\end{tabular}
	\caption{Major Manifold Learning Algorithms and their Assumptions.}	
\end{sidewaystable}

\subsection{Manifold Assumption}
One fundamental assumption of manifold learning is that the input data points lie on or nearly on a manifold $\mathcal{M}$ which is viewed as a sub-manifold of the ambient feature space. For each algorithm, it has additionally special assumptions. IsoMap assumes that $\mathcal{M}$ is globally isometric to a convex subset of Euclidean space.
Locally preserved manifold learning algorithms \cite{20} visualize the embedded manifold as a collection of overlapping local patches. For different local preserved manifold learning algorithms, the assumptions for local patches are different.
LLE assumes that $\mathcal{M}$ is an open sub-manifold and the input data points are dense enough to make the neighborhood of each data point a linear subspace. LEP also regards the neighborhood of each sample as a linear subspace, then constructs the corresponding local weight matrix, where the distance between two neighbor samples is measured by Euclidean metric. HLLE assumes that $\mathcal{M}$ is locally isometric to Euclidean space, so that the null space can be uncovered by the average norm of Hessian matrix of all data points. For LTSA, in each local patch it uses PCA \cite{12} to reduce the dimension of local samples. So it assumes that each local patch of $\mathcal{M}$ is a linear subspace of Euclidean space. PFE \cite{17} uses parallel vector field to learn a dimension reduction map, where this map induces $\mathcal{M}$ locally isometric to Euclidean space. LSML \cite{10} reduces the dimension of $\mathcal{M}$ which is not isometric to Euclidean space. But it regards the local patches of sub-manifold as linear subspaces. All the assumptions of MAL algorithms are shown in Table 1. 

All the MAL algorithms except for LSML assume that $\mathcal{M}$ is globally or locally isometric to Euclidean space. 
But in practice, general manifold is far-fetched to satisfy these assumptions. All the existing algorithms do not analyze the reliability and validity of these assumptions. In addition, all of them do not analyze the difference between non-isometry property and isometry property of manifold. 

\subsection{Limitations}

Despite the wide applications of the existing MAL algorithms in many fields, such as: computer vision, pattern recognition, and machine learning, there are still a few limitations and problems remained to be solved.

\begin{itemize}
	\item  \textbf{Local linearity assumption:} it requires the input data points to be dense enough to guarantee the local patches being linear subspaces. In practice, there are not enough samples to generate the local patches with small enough size to guarantee the linearity. 
	\item  \textbf{Parameters sensitivity problem:} the neighbor-size parameter determines the size of local patches. Since the local isometry hypothesis, it requires the neighbor-size small enough. Otherwise, it would break the assumption of existing manifold learning algorithms. 
	\item  \textbf{Locally short circuit problem:} if the embedded manifold is highly curved, the local Euclidean distance between any two points is obviously shorter than the intrinsic geodesic distance.
	\item   \textbf{Intrinsic dimension estimation problem: }since local patches are simply taken as tangent spaces, the intrinsic dimension of manifold cannot be determined by the latter accurately, in particular in case of strongly varying curvature.
	\item  \textbf{Curvature sensitivity problem:} if the curvature of original manifold is especially high at some point, smaller patch is needed for representing the neighborhood around this point. In practice, it is hard to avoid this case, especially when the data points are sparse.
\end{itemize}

All the limitations mentioned above generate from the assumption that $\mathcal{M}$ is locally isometric to Euclidean space. Thus, to remove this assumption is the main target of this paper. The problem of our paper is stated as follows. 

\subsection{Problem Statement}

The input data points that we considered in this paper are $\{x_1, x_2, \cdots, x_N\}\in \mathbb{R}^D$, where $N$ is the number of data points and $D$ is the dimension of data points. We assume that these discrete data points lie on a $d$-dimensional manifold $\mathcal{M}$ embedded in the high dimensional feature space $\mathbb{R}^D, d \ll D$, where $\mathcal{M}$ can be viewed as a sub-manifold of $\mathbb{R}^D$. The aim of manifold learning is to learn an embedding map $f$:
\begin{equation}
x_i = f\left(y_i \right)+ \epsilon_i, i=1,\cdots,N,
\end{equation}
where $\{y_1, \cdots, y_N\} \in \mathbb{R}^d$ are lower dimensional representations of $\{x_1,\cdots,x_N\}$ and $\{\epsilon_1, \epsilon_2, \cdots, \epsilon_N\}$ are the corresponding noises. $f$ needs to preserve the geometric structure of sub-manifold so that the lower dimensional representations can uncover the intrinsic structure of sub-manifold $\mathcal{M}$.

Under local isometry assumption, the embedding map $f$ locally satisfies:
\begin{equation}
\|f\left(y_i\right) - f\left(y_j\right)\|^2 = \|y_i-y_j\|^2 + o\left(\|y_i-y_j\|^2\right),
\end{equation}
where $y_i, y_j$ are in a same local patch.

For general manifold, the locally isometric condition is not always satisfied, such as sphere \cite{10}. The problem that we aim to solve in this paper is the situation that $\mathcal{M}$ is non-locally isometric to Euclidean space. All the manifold learning algorithms aim to uncover the intrinsic structure of the embedded-manifold $\mathcal{M}$. Thus our method attempts to learn the embedding map $f$ in Eq.1 under non-isometric condition which is not satisfied Eq.2. In the next section, we give a detailed analysis to make clear the relationship between local isometry and curvature tensor of sub-manifold $\mathcal{M}$. Based on this analysis, we give our curvature-aware manifold learning algorithm.

\section{Geometry Background}
In this section, we first give the definition of local isometry, then we give a geometric interpretation behind the locally isometric assumption. From this analysis, we uncover the potential limitations of traditional manifold learning algorithms. In second subsection, we give the geometry theory of general Riemannian sub-manifold.  

\subsection{Local Isometry}

The family of all inner products defined on all tangent spaces is known as Riemannian metric $g$ of manifold $\mathcal{M}$. At each tangent space $T_p \mathcal{M}$, the Riemannian metric is a scalar inner product $g_p$, $p \in \mathcal{M}$.

\textbf{Definition 2.1. (Local Isometry)} \cite{13} \textsl{Let $\left(\mathcal{M},g\right)$ and $\left(\mathcal{N},h\right)$ be two Riemannian manifolds where $g$ and $h$ are Riemannian metrics on them. For a map between manifolds $F : \mathcal{M} \rightarrow \mathcal{N}$, $F$ is called local isometry if $h \left(dF_p\left(v\right),dF_p\left(v\right)\right) = g \left(v,v\right)$ for all $p \in \mathcal{M}, v \in T_p \mathcal{M}$. Here $dF$ is the differential of $F$.}

Under local isometry, $dF$ is a linear isometry between the corresponding tangent spaces $T_p\mathcal{M}$ and $T_{F\left(p\right)} \mathcal{N}$.

\textbf{Definition 2.2. (Global Isometry)} \cite{13} \textsl{A map $F : \mathcal{M} \rightarrow \mathcal{N}$ is called global isometry between manifolds if it is a diffeomorphism and also a local isometry.}


A Riemannian manifold is said to be \textsl{flat} if it is locally isometric to Euclidean space. That is to say, if every point has a neighborhood isometric to an open subset of Euclidean space, the Riemannian manifold is called a flat manifold.

\textbf{Theorem 2.1.} \cite{14} \textsl{A Riemannian manifold is flat if and only if its curvature tensor vanishes identically}.

So under the local isometry assumption of traditional MAL, the curvature tensor of sub-manifold $\mathcal{M}$ is null tensor everywhere. However, in general the sub-manifold may be highly curved and not isometric to Euclidean space. Under this case, traditional manifold learning algorithms cannot accurately uncover the intrinsic structure of sub-manifold. 

The root cause of these limitations for traditional manifold learning algorithms is without considering the curvature tensor of sub-manifold. To our knowledge, there have been several papers to consider the intrinsic curvature of data points \cite{7} \cite{22} \cite{23}. However, K. I. Kim et al. \cite{7} mainly applied on semi-supervised learning. Xu et al. \cite{22} and \cite{23} used Ricci flow to rectify the pair-wise non-Euclidean dissimilarities among data points. In this paper, our method attempts to add curvature information in manifold learning and remove the limitations of traditional manifold learning. Thus we propose our curvature-aware manifold learning.

%

\subsection{Riemannian Sub-manifold}




In Riemannian geometry, the geometric structure of sub-manifold $\mathcal{M}$ is determined by two fundamental forms. Riemannian metric $g$ can be viewed as the \textsl{first fundamental form} which aims to compute the intrinsic geometric structure of Riemannian manifold, such as: the geodesic distance, area, and volume. The \textsl{second fundamental form} aims to uncover the extrinsic structure of sub-manifold $\mathcal{M}$ relative to ambient space, such as curvature, torsion and so on. For Riemannian manifold, the torsion is zero. 
How the sub-manifold $\mathcal{M}$ curved with respect to the ambient space is measured by the \textsl{second fundamental form}.

\subsection{Second fundamental form}
Suppose $\left(\widetilde{\mathcal{M}},\tilde{g}\right)$ is a Riemannian manifold with dimension $D$ and $\left(\mathcal{M},g\right)$ is embedded in $\left(\widetilde{\mathcal{M}},\tilde{g}\right)$ with dimension $d$. At any point $p \in \mathcal{M}$, the ambient tangent space $T_p \widetilde{\mathcal{M}}$ divides into two perpendicular linear subspaces $T_p \widetilde{\mathcal{M}} = T_p \mathcal{M} \oplus N_p \mathcal{M}$ \cite{14}, where $N_p \mathcal{M} \doteq \left(T_p \mathcal{M}\right)^\bot$ is the normal space and $T_p\left(\mathcal{M}\right)$ is the tangent space of $\mathcal{M}$ at $p$. In this paper, we regard Riemannian manifold $\mathcal{M}$ as a Riemannian sub-manifold of $\mathbb{R}^D$.
The Riemannian metric $g$ of $\mathcal{M}$ is defined as the induced metric from $\mathbb{R}^D$. 
Riemannian curvature tensor defined on Riemannian manifold is a $4^{th}$ order tensor. The curvature operator is represented by the second order derivative on vector field of Riemannian manifold, where the directional derivative is defined as Riemannian connection $\nabla$. In Riemannian sub-manifold, the Riemannian curvature tensor of sub-manifold is computed with the help of \textsl{second fundamental form} expressed as $\mathcal{B}$.

\textbf{Definition 2.3. (Riemannian Curvature)} \cite{15} \textsl{ Let $\left(\mathcal{M},g\right)$ be a Riemannian manifold and $\nabla$ the Riemannian connection. The curvature tensor is a $\left(1,3\right)$-tensor defined by:}
\begin{equation*}
\mathcal{R} \left(X,Y\right)Z=\nabla_X \nabla_Y Z - \nabla_Y \nabla_X Z-\nabla_{[X,Y]} Z,
\end{equation*}
on vector fields $X,Y,Z$. 

Using Riemannian metric $g$, $\mathcal{R} \left(X,Y\right)Z$ can be changed to a $\left(0,4\right)$-tensor \cite{15}:
\begin{equation}
\mathcal{R}\left(X,Y,Z,W\right)=g\left(\mathcal{R}\left(X,Y\right)Z,W\right).
\end{equation}
In Riemannian sub-manifold, one main task is to compare the Riemannian curvature of $\mathcal{M}$ with that of ambient space $\widetilde{\mathcal{M}}$. According to the definition of curvature tensor, we first give the relationship between the Riemannian connection $\nabla$ of $\mathcal{M}$ and $\widetilde{\nabla}$ of $\widetilde{\mathcal{M}}$ \cite{14}:
\begin{equation}
\widetilde{\nabla}_X Y = \nabla_X Y + \mathcal{B}\left(X,Y\right),
\end{equation}
where the normal component is known as the \textsl{second fundamental form} $\mathcal{B}\left(X,Y\right)$ of $\mathcal{M}$.

Therefore, we can interpret the \textsl{second fundamental form} as a measure of the difference between the Riemannian connection on $\mathcal{M}$ and the ambient Riemannian connection on $\widetilde{\mathcal{M}}$. 
Based on the relationship between $\nabla$ and $\widetilde{\nabla}$, we give the following theorem to show the relationship between the Riemannian curvature of sub-manifold and the Riemannian curvature of ambient space.

\textbf{Theorem 2.2. (The Gauss Equation)} \cite{14} \textsl{For any vector fields $X,Y,Z,W \in \mathcal{T} \left(\mathcal{M}\right)$, the following equation holds:}
\begin{equation*}
\begin{split}
&\widetilde{\mathcal{R}}\left(X,Y,Z,M\right) = \mathcal{R} \left(X,Y,Z,W\right)-\\&\langle B\left(X,W\right),B\left(Y,Z\right)\rangle + \langle B\left(X,Z\right), B\left(Y,W\right)\rangle.
\end{split}
\end{equation*}

So Riemannian curvature of ambient space can be decomposed into two components. In this paper the ambient space is Euclidean space $\mathbb{R}^D$, so $\widetilde{\mathcal{R}}\left(X,Y,Z,W\right)=0$. In this case, the Riemannian curvature of $\mathcal{M}$ is represented as:
\begin{equation}
\begin{split}
& \mathcal{R} \left(X,Y,Z,W\right) = \langle \mathcal{B}\left(X,W\right), \mathcal{B}\left(Y,Z\right) \rangle \\& - \langle \mathcal{B}\left(X,Z\right), \mathcal{B} \left(Y,W\right) \rangle.
\end{split}
\end{equation}
In order to compute the scalar value of \textsl{second fundamental form}, we construct a local natural orthonormal coordinate frame $\{\frac{\partial}{\partial x^1}, \cdots, \frac{\partial}{\partial x^d}, \frac{\partial}{\partial y^1}, \cdots,\frac{\partial}{\partial y^{D-d}}\}$ of the ambient space $\widetilde{\mathcal{M}}$ at point $p$, the restrictions of $\{\frac{\partial}{\partial x^1}, \cdots, \frac{\partial}{\partial x^d}\}$ to $\mathcal{M}$ form a local orthonormal frame of $\mathcal{T}_p\left(\mathcal{M}\right)$. The last $D-d$ orthonormal coordinates $\{\frac{\partial}{\partial y^1}, \cdots,\frac{\partial}{\partial y^{D-d}}\}$ form a local orthonormal frame of $\mathcal{N}_p \left(\mathcal{M}\right)$. Under the locally natural orthonormal coordinate frame, the Riemannian curvature of $\mathcal{M}$ in Eq.5 is represented as:
\begin{equation}
R_{jkl}^i =\sum_{\alpha} \left(h_{ik}^\alpha h_{jl}^\alpha - h_{il}^\alpha h_{jk}^\alpha\right).
\end{equation} 
Accordingly, the \textsl{second fundamental form} $\mathcal{B}$ under this local coordinate frame is showed as:
$\mathcal{B}\left(\frac{\partial}{\partial x^i},\frac{\partial}{\partial x^j}\right) = \sum_{\alpha=1}^{D-d}h^\alpha_{ij} \frac{\partial}{\partial y^\alpha}$, 
with $h^\alpha_{ij} \left(\alpha=1, \cdots, D-d\right)$ being the coefficients of $\mathcal{B}\left(\frac{\partial}{\partial x^i},\frac{\partial}{\partial x^j}\right)$ with respect to the normal coordinate frame $\{\frac{\partial}{\partial y^1}, \cdots,\frac{\partial}{\partial y^{D-d}}\}$.  And under this locally natural coordinate frame, the embedding map $f$ is redefined as $f(x^1,x^2,\cdots,x^d )=[x^1,x^2,\cdots,x^d,f^{1},\cdots,f^{D-d}]$, where $x \doteq [x^1,x^2, \cdots,x^d]$ are natural parameters.
$h_{ik}^\alpha$ is the second derivative $\frac{\partial^2f^\alpha}{\partial x^i \partial x^j}$ of embedding component function $f^\alpha$, which constitutes the Hessian matrix $H^\alpha=\left(\frac{\partial^2 f^\alpha}{\partial x^i \partial x^j}\right)$. By the above analysis, in order to compute the Riemannian curvature of Riemannian sub-manifold $\mathcal{M}$, we just need to estimate the Hessian matrix of the embedding map $f$. Next, we give the estimation of Hessian operator.

\subsection{Hessian Operator}

Hessian matrix is a square matrix of second-order derivatives with respect to all of the variables of a scalar-valued function. It represents the concavity, convexity and the local curvature of a function. Suppose  $f^\alpha(x^1,x^2,\cdots,x^d)$ is a multivariate function with $d$ parameters. Then the Hessian matrix $H$ of $f^\alpha$ is given as: $H_{ij} (f^\alpha)=\frac{\partial^2 f^\alpha}{\partial x^i \partial x^j}$.

In each local patch $U_i$ of $x_i$, we choose a set of local natural orthogonal coordinates $\{\frac{\partial}{\partial x^1}, \frac{\partial}{\partial x^2},\cdots, \frac{\partial}{\partial x^d}\}$. In practice, we use PCA \cite{12} to estimate the local orthogonal coordinate system of $\mathcal{M}$. The corresponding normal coordinates of normal space are computed by \textsl{Gram-Schmidt orthogonal} method. The corresponding local coordinates of $\{x_{i_1},\cdots, x_{i_K}\}\in U_i$ under this new local normal coordinate system are represented as $\{u_{i_1},u_{i_2},\cdots,u_{i_K}\}$. $x_i$ is projected into the original point. The \textsl{second fundamental form} coefficients are estimated as $h_{ij}^\alpha=\frac{\partial^2 f^\alpha}{\partial x^i \partial x^j},\alpha=1,\cdots,D-d$.

Consider the Taylor expansion of $f^\alpha, \left(\alpha=1, \cdots, D-d\right)$, at $x_i$ under this new local coordinate system:
\begin{equation}
f^\alpha \left(u_{i_j}\right) = f^\alpha \left(0\right) + u_{i_j} \nabla f^\alpha + u_{i_j} H^\alpha u_{i_j}^T + o\left(\|u_{i_j}\|^2\right).
\end{equation}

For each component $h_{ij}^\alpha = \frac{\partial^2 f^\alpha}{\partial x^i \partial x^j}$ of Hessian matrix $H^\alpha$, it can be considered as the second order item coefficient of the quadratic polynomial function $f^\alpha$. For local tangent space at $x_i$, the orthogonal coordinate system is spanned by $\{\frac{\partial}{\partial x^1}, \frac{\partial}{\partial x^2}, \cdots, \frac{\partial}{\partial x^d}\}$. For local quadratic polynomial vector space, the local coordinate system is spanned by $\{\frac{\partial}{\partial x^1}, \cdots, \frac{\partial}{\partial x^d},\frac{\partial^2}{\partial^2 x^1},\cdots, \frac{\partial^2}{\partial^2 x^d}, \frac{\partial^2}{\partial x^1 \partial x^2}, \cdots, \frac{\partial^2}{\partial x^{d-1} \partial x^d}\}$. So the Hessian matrix $H^\alpha$ is estimated by projecting the input data points into the polynomial vector space. For estimation, we use the \textsl{least square estimation} method to compute the projecting coefficents. The solution is obtained by:
$B_i=\Psi^\dagger f$,
where $\Psi_{i_j}=[1, u_{i_j}^1, \cdots, u_{i_j}^d,$ $ \left(u_{i_j}^1\right)^2,\cdots,\left(u_{i_j}^d\right)^2,\left(u_{i_j}^1 \times u_{i_j}^2\right), \cdots, \left(u_{i_j}^{d-1} \times u_{i_j}^d\right)]$, $\Psi=[\Psi_{i_1}, \cdots, \Psi_{i_K}]$, $\Psi^\dagger$ is the pseudo-inverse matrix of $\Psi$ and $f=[f^1,f^2,\cdots, f^{D-d}]$, $f^\alpha=[f^\alpha\left(u_{i_1}\right),\cdots,$ $ f^\alpha\left(u_{i_K}\right)]^T, \alpha=1,2,\cdots,D-d$.
The learnt local projection coordinates of each point $x_{i_j} \in U_i$ is given as $B_{i_j}=[1, \tau_{i_j}, H^{i_j}]$ where $\tau_{i_j}$ is the tangent components vector and $H^{i_j}$ is the vector-form representation of Hessian matrix. $x_i$ is projected into the original point expressed as $B_{i_0}=0$.

\section{Curvature-aware Manifold Learning}

In this paper, we just consider the locally geometric structure preserving MAL algorithms, namely LLE, LEP, LTSA and so on.
These algorithms attempt to recover the local underlying structure of sub-manifold $\mathcal{M}$ in lower dimensional Euclidean space. In general, the procedures of this type of algorithms are mainly divides into three steps \cite{8}.
The detailed statement is given in the following subsection.

\subsection{Manifold Learning}

In the first step, traditional MAL algorithms partition local patches $\{U_i\}$ to each input point $x_i$ based on the Euclidean metric in ambient space $\mathbb{R}^D$. In general, there are two commonly used methods. The first one is by choosing an $\varepsilon$-ball with $x_i$ as center. Then all the points in this ball are called the neighbors of $x_i$. The other method is to use $K$-nearest neighbor method to find the neighbors of each input data point $x_i$. 
For these two methods, $\varepsilon$ and $K$ are parameters which are very sensitive to the dimension reduction results of experiments.

In the second step, traditional manifold learning algorithms aim to construct a weight matrix $W_i$ in each local patch $U_i$ to represent the local geometric structure of sub-manifold $\mathcal{M}$. For different manifold learning algorithms, the weight matrices are different. 


The third step is to reconstruct a set of lower dimensional representations $Y=\{y_1,\cdots,y_N\}$, where $y_i \in \mathbb{R}^d$ corresponds to $x_i$. $Y$ is learnt by minimizing a reconstruction error function $\Phi$ under some normalization constraints \cite{8}. 
\begin{equation}
\Phi \left(Y\right)=\sum_{i=1}^{N} \phi \left(Y_i\right)=\sum_{i=1}^{N} \|W_iY_i\|_F^2,
\end{equation}
with the normalization constraints $Y^T Y =I, Y'1=0$ for LLE, LTSA, HLLE and $Y^TDY=I, Y^TD1=0$ for LEP where $D$ is a diagonal matrix with $D_{ii}= \sum_{j=1}^{N}W_{ij}$.


\subsection{Curvature-aware Manifold Learning}

It has been analyzed that one critical assumption of traditional manifold learning algorithms is that the embedded manifold is isometric to Euclidean space. For this type of algorithms, the similarity between any two neighbor points is measured by Euclidean distance. Obviously, it overestimates the similarity if the manifold is highly curved. 

In LTSA \cite{5}, the authors analyzed the reconstruction error in theory and obtained that the error is highly influenced by the curvature of sub-manifold $\mathcal{M}$. When the sub-manifold is highly curved in the higher dimensional feature space, the reconstruction error would be very high. By analysis, the accurate determination of local tangent space is dependent on several factors: curvature information embedded in the Hessian matrices, local sampling density, and noise level of data points. So for LTSA, it is necessary to analyze the curvature information of sub-manifold during dimension reduction.


Besides analyzing the reconstruction error of LTSA, our method aims to improve traditional manifold learning algorithms by adding curvature information. In this paper, we focus on improving two algorithms LLE and LEP in detail and give the detailed analysis of these two improved algorithms CA-LLE and CA-LEP in theory. For local structure preserved method, we just divide the sub-manifold $\mathcal{M}$  into a set of local patches $\{U_i\}$ in each point $x_i$ and choose $U_i$ as an example to analyze our algorithm.  
We have shown that we consider the local patch structure in quadratic polynomial vector space to obtain curvature information. In local patch $U_i$ of $x_i$, $\{\frac{\partial}{\partial x^1}, \cdots, \frac{\partial}{\partial x^d}, \frac{\partial^2}{\partial^2 x^1}, \cdots, \frac{\partial^2}{ \partial^2 x^d}, \frac{\partial^2}{\partial x^1 \partial x^2}, \cdots, \frac{\partial^2}{\partial x^{d-1} \partial x^d}\}$ span the local polynomial vector space. Projecting original input data points $x_{i_j} \in U_i$ to this local polynomial vector space, we respectively obtain the corresponding projection coefficients shown as $B_{i_j}=[1, \tau_{i_j}, H^{i_j}]$. 
The local curvature information of $U_i$ at $x_{i_j}$ is hidden in the quadratic component vector $H^{i_j}$. In the following, we give the detailed description of our CAML algorithm.
\\
\\
\textbf{CAML Algorithm Procedures:}
\\
\\
1. 
\begin{minipage}[t]{0.95\linewidth}
	Input a set of data points $x_1,x_2\cdots,x_N \in \mathbb{R}^D$. This step is the same as the first step of traditional manifold learning algorithm. 
	In this step, we choose $K$ \textsl{nearest neighbor} method to divide the sub-manifold $\mathcal{M}$ into a set of local patches $\{U_1, U_2, \cdots, U_N\}$ in each point $\{x_i\}$ under Euclidean metric.\\
\end{minipage}\\
2.
\begin{minipage}[t]{0.95\linewidth}
	Unlike traditional manifold learning algorithms to project local patches to local tangent spaces, we project the local patch $U_i$ into a second-order polynomial vector space and obtain the new local coordinate representations $\{B_{i_1}, B_{i_2}, \cdots, B_{i_K}\}$ by Eq.7, where $B_{i_j}=[1,\tau_{i_j}, H^{i_j}]$. The curvature information at each point $x_{i_j}$ is embedded in $H^{i_j}$.\\
\end{minipage}\\
3.
\begin{minipage}[t]{0.95\linewidth}
	Using new local representations to construct the local geometrical structure represented by weight matrix $W_i$ in each local patch $U_i$. 
	This step is the most critical step of CAML by adding curvature information to reconstruct the local weight matrix $W$.\\
\end{minipage}\\
4.
\begin{minipage}[t]{0.95\linewidth}
	After constructing the curvature-aware weight matrix $W$, we use this weight matrix to reconstruct the representations $Y=\{y_1,y_2,\cdots,y_N\}$ in lower dimensional Euclidean space $\mathbb{R}^d$. $Y$ is learnt by minimizing a reconstruction error function in Eq.8 under some normalization constraints. This step is the same as the third step of traditional manifold learning.
\end{minipage}\\

We respectively consider the improvements of LEP and LLE under our curvature-aware algorithms as two examples:\\
\\
\textbf{For Curvature-aware LEP:}
\begin{equation}
W_{ij}=\left\{\begin{array}{lr}
exp^{-\frac{\|B_{i_0}-B_{i_j}\|^2}{2\sigma^2}}, x_{i_j} \in U_i & \\
0, x_{i_j} \notin U_i &,
\end{array}
\right.
\end{equation}\\
\textbf{For Curvature-aware LLE:}
\\
$\{W_{ij}\}$ in local patch $U_i$ is obtained by minimizing the following equation:
\begin{equation}
arg min \|B_{i_0} - W_{ij} B_{i_j}\|^2.
\end{equation}
When we use $B_{i_j}$ to represent the local coordinate representation of $x_{i_j}$, the curvature information of local patches is added into the local weight matrix $W_i$. The detailed theoretical analysis is shown in the following section.

\begin{algorithm}[tb]
	\caption{Curvature-aware Manifold Learning}
	\label{alg:example}
	\begin{algorithmic}
		\STATE {\bfseries Input:} Training data points $\{x_1,x_2,\cdots ,x_N\}\in \mathbb{R}^D$, neighbor size parameter $K$.\\
		\STATE {\bfseries Output:} $\{y_1,y_2,\cdots y_N\} \in \mathbb{R}^d$\\
		1.
		\begin{minipage}[t]{0.9\linewidth}
			\textbf{for} $i=1$ \textbf{to} $N$ \textbf{do} 
		\end{minipage}\\
		2.
		\begin{minipage}[t]{0.9\linewidth}
			\qquad Find $K$-nearest neighbors of $x_i$;
		\end{minipage}\\
		3.
		\begin{minipage}[t]{0.9\linewidth}
			\textbf{end for}
		\end{minipage}\\
		4.
		\begin{minipage}[t]{0.9\linewidth}
			Determine the intrinsic dimension $d$ of $\mathcal{M}$.
		\end{minipage}\\
		5.
		\begin{minipage}[t]{0.9\linewidth}
			\textbf{for} $i=1$ \textbf{to} $N$ \textbf{do} 
		\end{minipage}\\
		6.
		\begin{minipage}[t]{0.9\linewidth}
			\qquad Compute vector space $B_i$ by Eq.7.
		\end{minipage}\\
		7.
		\begin{minipage}[t]{0.9\linewidth}
			\qquad Construct the local weight matrix $W_i$.
		\end{minipage}\\	
		8.
		\begin{minipage}[t]{0.9\linewidth}
			\qquad Eq.9 for CA-LEP; Eq.10 for CA-LLE. 
		\end{minipage}\\
		9.
		\begin{minipage}[t]{0.9\linewidth}
			\textbf{end for}
		\end{minipage}\\			 		 
		10.
		\begin{minipage}[t]{0.9\linewidth}
			Minimize the reconstruction error function Eq.8.
		\end{minipage}	  		
	\end{algorithmic}
\end{algorithm}

\section{Algorithm Analysis}

We just consider one local patch $U_i$ as an example to give the analysis of our local curvature-aware manifold learning algorithm. Different from traditional locally preserved MAL algorithms, our method projects the original data points in local polynomial vector space. The corresponding local projection of $x_{i_j} \in U_i$ is shown as $B_{i_j}=[1,\tau_{i_j},H^{i_j}]$. 

\subsection{Curvature-aware LEP}

In the polynomial vector space, the weight value $W_{ij}$ between two neighbor points is given as:
\begin{equation}
W_{ij}=exp^{-\frac{\|B_{i_0}-B_{i_j}\|^2}{2\sigma^2}}=exp^{-\frac{\|\tau_{i_0}-\tau_{i_j}\|^2}{2\sigma^2}}\cdot exp^{-\frac{\|H^{i_0}-H^{i_j}\|^2}{2\sigma^2}},
\end{equation}
where $\|H^{i_0}-H^{i_j}\|^2 = \|H^{x_j}\|_F^2$. $H^{x_j}$ represents the Hessian matrix at point $x_j$. Under this new local normal coordinate frame of $U_i$, the coordinate of $x_i$ is zero, so $H^{i_0}=0$, obviously $\|H^{i_j}\|^2=\|H^{x_j}\|_F^2$.

Hessian matrix is a symmetric matrix. We do the eigenvalue decomposition to $H^{x_j}$ and obtain the following expression:
\begin{equation}
\|H^{x_j}\|_F^2 =\|U^T \Lambda_j U\|_F^2 = \|\Lambda_j\|_F^2,
\end{equation}
where $\Lambda_j$ is the eigenvalue matrix of $H^j$. In Riemannian geometry, each eigenvalue of $H$ is a principal curvature along the corresponding coordinate.
Based on the above analysis, the weight value $W_{ij}$ in Eq.11 is shown as:
\begin{equation}
W_{ij} = exp^{-\frac{\|\tau_{i_0}-\tau_{i_j}\|^2}{2\sigma^2}}\cdot exp^{-\frac{\|\Lambda_j\|_F^2}{2\sigma^2}}.
\end{equation}
It is equivalent to add a curvature penalty on similarity weight $W$.

\textbf{Theorem 4.1.}\textsl{ Assume the reconstruction error under our curvature-aware weight matrix is represented as $E$. And the reconstruction error under traditional LEP \cite{1} is represented as $\widetilde{E}$. Then, we have}
\begin{equation}
\|E\|_F \leq \|\widetilde{E}\|_F.
\end{equation}
\\
\textbf{Proof:} The weight matrix $W$ under our CAML is defined as in Eq.13: 
\begin{equation*}	W_{ij} = exp^{-\frac{\|\tau_{i_0}-\tau_{i_j}\|^2}{2\sigma^2}}\cdot exp^{-\frac{\|\Lambda_j\|_F^2}{2\sigma^2}}.
\end{equation*}
And the weight $\widetilde{W}$ under traditional LEP algorithm is defined as:
\begin{equation*}
\widetilde{W}_{ij} = exp^{-\frac{\|\tau_{i_0}-\tau_{i_j}\|^2}{2\sigma^2}}.
\end{equation*}
Obviously we have $W_{ij}\leq \widetilde{W}_{ij}$. The corresponding Laplace matrices are defined as $L=D-W$, $\widetilde{L}=\widetilde{D}-\widetilde{W}$. Therefore, we have:
\begin{equation*}
\lambda_i\left(L\right) \leq \lambda_i\left(\widetilde{L}\right), i=1,2,\cdots,N,
\end{equation*}
where $N$ is the number of input data points.

For LEP, the lower dimensional representations are obtained from the $d$ eigenvectors of the smallest $d$ eigenvalues of Laplace matrix. The reconstruction error $E$ is measured by the values of the smallest $d$ eigenvalues $\lambda_i, i=1,2,\cdots,d$,
\begin{equation*}
\|E\|_F=\sum_{i=1}^{N}\|x_i-f\left(y_i\right)\|=\sum_{i=1}^{d}\lambda_i.
\end{equation*}
We have proved that the eigenvalue of $L$ is less than that of $\widetilde{L}$. So we have 
\begin{equation*}
\|E\|_F \leq \|\widetilde{E}\|_F.
\end{equation*}
Therefore, when considering the curvature information of sub-manifold, the reconstruction error gets much lower. $\square$ 
	
\subsection{Curvature-aware LLE}
	
In each local patch $U_i$, we compute the local linear combination structure by minimizing the following equation:
\begin{equation}
\Phi_i=\|B_{i_0}-\sum_{j=1}^{K}W_{ij}B_{i_j}\|^2,
\end{equation}
where $B_{i_j}=[1, \tau_{i_j}, H^{i_j}],\sum_{j=1}^{K} W_{ij}=1$.
	
The equation $\Phi_i$ in Eq.15 can be rewritten as:
\begin{equation}
\Phi_i=\|\tau_{i_0}-\sum_{j=1}^{K}W_{ij}\tau_{i_j}\|^2+\|H^{i_0}-\sum_{j=1}^{K}W_{ij}H^{i_j}\|^2.
\end{equation}
For traditional LLE, the authors just minimized the first item of $\Phi_i$. For our method we add an item to measure the linear combination of Hessian matrices. 
	
In the following, we give a theoretical derivation to explain the necessity for adding the second Hessian item of $\Phi_i$.
	
Frist we give the Taylor expansion of embedding map $f$ in local patch $U_i$:
\begin{equation}
f\left(u\right) = f\left(0\right) + u^T\nabla f + \frac{1}{2}\left(u^T H u\right)+o\left(\|u\|^2\right),
\end{equation}
Under the Taylor expansion of $f$, we obtain the linear relationship between $x_i$ and the rest neighbors:
\begin{equation}
\begin{split}
f\left(0\right) - \sum_{j} W_{ij} f \left(u_{i_j}\right) &\approx f\left(0\right) -\sum_{j} W_{ij} f\left(0\right)\\ &-\sum_{j} W_{ij} u_{i_j}^T \nabla f - \frac{1}{2}\sum_{i}W_{ij} u_{i_j}^T H u_{i_j}.
\end{split}
\end{equation}
Since $\sum_{j} W_{ij} =1, u_{i_0}=0, \sum_{j} W_{ij} u_{i_j}=0,$
\begin{equation}
f\left(0\right)-\sum_{j} W_{ij} f\left(u_{i_j}\right) \approx -\frac{1}{2} \sum_{j}W_{ij} u_{i_j}^T H u_{i_j},
\end{equation}
where $\sum_{j}W_{ij} u_{i_j}^T H u_{i_j} = \sum_{j} W_{ij} H^{i_j}$.
	
We have stated that the coordinate of $x_i$ under this local normal coordinate frame is zero, so the corresponding Hessian matrix $H^{i_0}=0$. Therefore Eq.19 can be given as:
\begin{equation}
f\left(0\right)-\sum_{j} W_{ij} f\left(u_{i_j}\right) \approx \frac{1}{2}H^{i_0} - \frac{1}{2} \sum_{j} W_{ij} H^{i_j}.
\end{equation}
So for our method, it is necessary to add a Hessian item when constructing the local linear combination structure. Traditional LLE algorithm does this linear combination in local tangent space, while our method does this in local polynomial vector space to consider the local curvature information of $U_i$.
	
\subsection{Time Complexity Analysis}
	
In this subsection, we give the time complexity analysis of our algorithm compared with traditional manifold learning algorithms based on the number of data points $N$, the input dimension $D$, the intrinsic dimension $d$. Comparing with traditional manifold learning algorithms, the added time cost of our algorithm mainly focuses on the computation of Riemannian curvature information of databases. The main process of this step is to estimate the local analytical structure by fitting a two-order polynomial function in Eq.7. We get the Riemannian curvature of each local patch by computing the eigenvalues of each Hessian matrix, where the size of each Hessian matrix is $d \times d$, so the time cost of eigenvalue decomposition of each Hessian matrix is $O\left(d^2\right)$, the total time cost of the full samples is shown as $O\left(Nd^2\right)$. In general, the intrinsic dimension $d$ is far less than the input dimension $D$. In addition, only the time cost of finding $K$ nearest neighbors of all samples is $O\left(N^2 \left(D+K\right)\right)$.
	
In short, compared with the total time cost of traditional manifold learning algorithms, the total time cost of our algorithm CAML is slightly higher than them. If the number of samples $N$ is especially large, the added time cost of CAML can be ignored.
	
\section{Experiments}
In this section, we compare our algorithm CAML with several traditional MAL algorithms on four synthetic databases e.g. \textsl{Swiss Roll, Punctured Sphere, Gaussian}, and \textsl{Twin Peaks} \cite{16} as well as two real world data sets. For synthetic databases, we respectively learn the effectiveness of our algorithm on two tasks: dimension reduction and parameter sensitivity analysis. For real world data sets, we compare the classification performance of our algorithm with traditional MAL algorithms.
	
\subsection{Topology Structure}
	
Before dimension reduction, we first analyze the topology structures of the four synthetic databases. 
All the databases are generated from Matlab code 'mani.m' \cite{16}. For each database, it contains $2000$ points distributed on the corresponding synthetic manifold. Swiss Roll is a locally flat manifold which is locally isometric to Euclidean space.
For this data set, traditional manifold learning algorithms can uncover the intrinsic structure of Swiss roll accurately. For punctured sphere data set, these $2000$ data points lie on a two dimensional sphere which is embedded in $\mathbb{R}^3$. 
The curvature of this sphere is non-zero everywhere, so it is not locally or globally isometric to Euclidean space. Twin peaks manifold is a highly curved two-dimensional manifold embedded in three dimensional Euclidean space. It is not locally isometric to Euclidean space, so traditional manifold learning algorithms cannot accurately uncover the intrinsic structure of this curved synthetic manifold. Two dimensional Gaussian manifold is not also isometric to Euclidean space, where the Gauss curvature of Gaussian manifold is not zero everywhere.
	
Based on the analysis of these four synthetic manifolds, we compare our curvature-aware manifold learning algorithm with other traditional manifold learning algorithms in the next two subsection to emphasize the need for considering curvature information.
	
\subsection{Dimension Reduction}
	
In this subsection, we compare our algorithm CAML with other MAL algorithms on these four datasets. 
To evaluate the performance of our curvature aware algorithm, we compare our method with seven traditional MAL algorithms (e.g. MDS, PCA, IsoMap, LLE, LEP, DFM, LTSA).
The objective of this comparison is to map each dataset to two dimensional Euclidean space and then to analyze the neighborhood preserving ratio (NPR) \cite{18} of different algorithms. Table 2 shows the comparison results, where the neighbor-size parameter $K=10$. The neighborhood preserving ratio (NPR) is defined as:
\begin{equation}
NB = \frac{1}{KN}\sum_{i=1}^{N} |\mathcal{N}\left(x_i\right) \bigcap \mathcal{N}\left(y_i\right)|,
\end{equation}
where $\mathcal{N}\left(x_i\right)$ is the set of $K$-nearest sample subscripts of $x_i$, and $\mathcal{N}\left(y_i\right)$ is the set of $K$-nearest sample subscripts of $y_i$. $|\cdot|$ represents the number of intersection points.
	
Table 2 shows that for all but Swiss Roll dataset, the NPRs of our CA-LEP and CA-LLE are higher than the rest traditional MAL algorithms. Swiss Roll is a flat Riemannian manifold, so our algorithm has almost no advantages under this dataset. For Gaussian dataset, it is a symmetric and convex manifold. So the NPRs of all algorithms are all very high. For Punctured Sphere and Twin Peaks, the NPRs of our algorithm obviously outperform the rest traditional MAL algorithms. These results clearly demonstrate that our CAML algorithm is more stable and better to uncover the local structure of data points.
\begin{table*}[tbp]
	\centering  
	\begin{tabular}{|l|c|c|c|c|}  
		\hline
		Methods   &Twin Peaks &Swiss Roll &Punctured Sphere & Gaussian \\ \hline
		MDS \cite{11} &$0.4968$ &$0.4352$ &$0.5474$ &$0.9082$ \\ \hline 
		PCA \cite{12} &$0.4867$ &$0.4167$ &$0.3741$ &$0.8960$ \\ \hline
		LEP \cite{1} &$0.6541$ &$0.2145$ &$0.6449$ &$0.5400$ \\ \hline
		LLE \cite{6} &$0.7852$ &$0.6156$ &$0.6662$ &$0.8912$ \\\hline
		IsoMap \cite{3} &$0.7659$ &$0.7957$ &$0.5516$ &$0.8568$\\ \hline
		LTSA \cite{5} & $0.7748$ &$0.5143$ &$0.3893$ &$0.8960$ \\ \hline
		Diffusion Map \cite{9} &$0.4866$ &$0.2290$ &$0.4063$ &$0.8962$ \\ \hline
		CA-LEP &$\textbf{0.8016}$ &$0.2739$ &$\textbf{0.7736}$ &$\textbf{0.9002}$ \\ \hline
		CA-LLE &$\textbf{0.8065}$ &$0.6604$ &$\textbf{0.7340}$ &$\textbf{0.9475}$ \\ \hline
	\end{tabular}
	\caption{The Neighborhood Preserving Ratio (NPR) comparisons of our CA-LEP and CA-LLE algorithms with other seven traditional manifold learning algorithms under four datasets (Twin Peaks, Swiss Roll, Punctured Sphere, and Gaussian), and Neighbor-size parameter $K=10$.}
\end{table*} 
	
	\begin{figure*} \centering    
		\subfigure[Twin Peaks] { \label{fig:a}     
			\includegraphics[width=0.48\columnwidth]{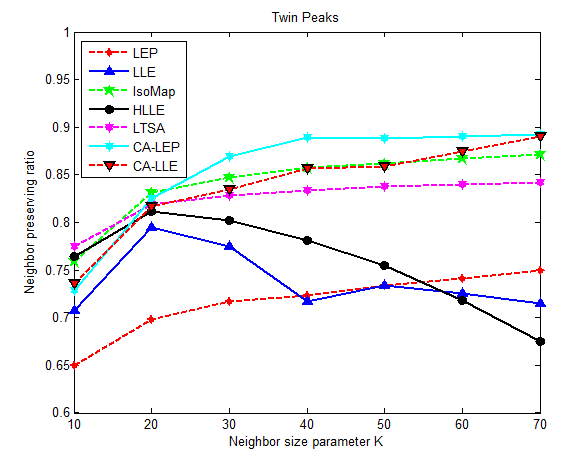}  
		}     
		\subfigure[Puncture Sphere] { \label{fig:b}     
			\includegraphics[width=0.45\columnwidth]{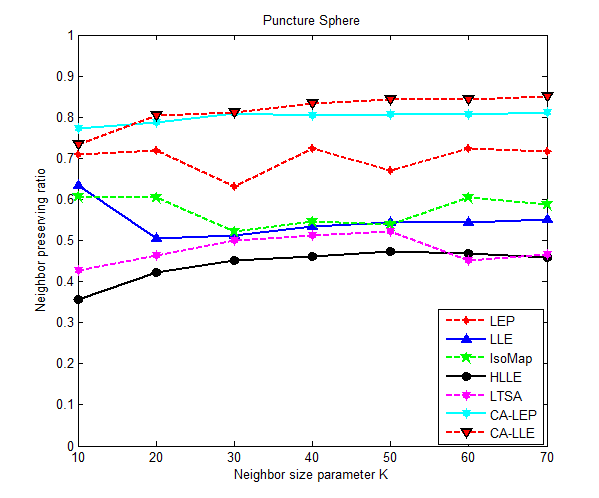}     
		}     
		\caption{Neighbor-size parameter sensitivity analysis about Twin Peaks dataset (a) and Puncture Sphere dataset (b). We choose $K=10,20,30,40,50,60,70$ respectively. And we compare our CA-LEP and CA-LLE algorithms with five traditional manifold learning algorithms (LEP, LLE, IsoMap, HLLE, and LTSA). }     
		\label{fig}     
	\end{figure*}

\subsection{Parameter Sensitivity Analysis}
By analyzing, traditional MAL algorithms are sensitive to some parameters e.g. neighbor-size parameter $K$, intrinsic dimension $d$. For intrinsic dimension $d$, we have not a very suitable method to estimate it exactly. So, in this paper, we assume that the intrinsic dimension $d$ of sub-manifold $\mathcal{M}$ is unique and approximately estimated \cite{19}. In this experiment, we mainly analyze the sensitivity of neighbor-size parameter $K$. 
	
We compare the neighborhood preserving ratios of different manifold learning algorithms under different parameter values $K$, $\left(K=10, 20, 30, 40, 50, 60, 70\right)$ respectively. All the experiments are done on two datasets (Punctured Sphere and Twin Peaks) with 2000 data points. These two synthetic manifolds are all not isometric to Euclidean space. Thus we analyze the effectiveness of curvature information in dimension reduction. In order to highlight the improvement of our algorithm, we use five traditional manifold learning algorithms (LEP, IsoMap, LLE, HLLE, and LTSA) to compare with our algorithm CAML. The final comparison results are shown in Figure 1. Compared with these traditional MAL algorithms, our method outperforms them when $K\geq 20$. In addition, from figure (a) and (b) we can see that the neighbor-size parameter $K$ is very sensitive under traditional MAL algorithms. The NPR curves under traditional MAL algorithms are changed especially unsteadily under different values of neighbor-size parameter $K$. However, the NPRs of CA-LEP and CA-LLE are steady growth as the increase of neighbor-size parameter $K$.
	
\subsection{Real World Experiments}
	
In this experiment, we consider the application of our algorithm on two real-world data sets: Extended Yale Face database B and USPS database. The main purpose of this experiment is to test the classfication accuracies in the lower dimensional space after using manifold learning algorithms to reduce the dimension of data points.
	
The Extended YaleFace B database, or YFB DB for short, contains $2414$ single light source images of $38$ individuals each seen under about $64$ near frontal images under different illuminations per individual. For every subject in a particular pose, an image with ambient illumination was also captured. The face region in each image is resized into $32 \times 32$, so the original dimension of this database is $1024$. 

\begin{figure*} \centering    
	\subfigure[YaleB database] { \label{fig:a}     
		\includegraphics[width=0.45\columnwidth]{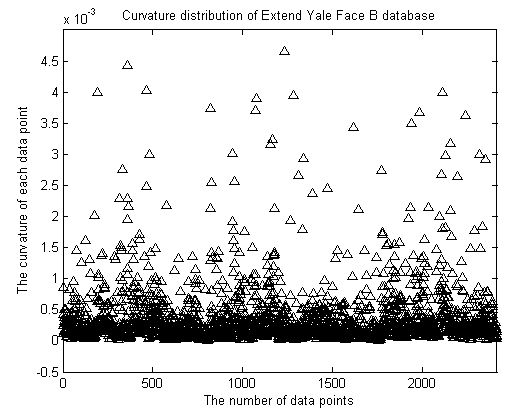}  
	}     
	\subfigure[USPS database] { \label{fig:b}     
		\includegraphics[width=0.465\columnwidth]{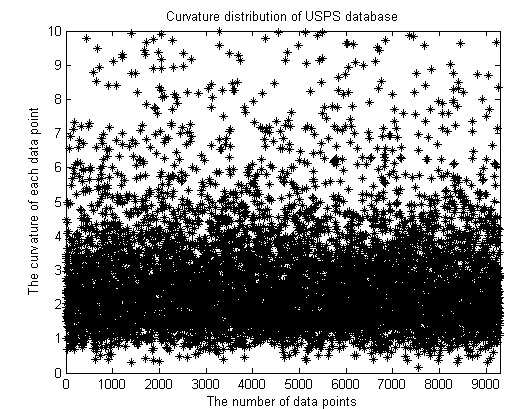}     
	}     
	\caption{The curvature distributions of Yale face B database (a) and USPS database (b).}     
	\label{fig}     
\end{figure*}
	
The USPS database consists of $9298$ images. It refers to numeric data obtained from the scanning of handwritten digits from envelopes by the U.S. Postal Service. The original scanned digits are binary and of different sizes and orientations; the images here have been deslanted and size normalized, resulting in $16 \times 16$ grayscale images. So the original dimension of this database is $256$.
	
In this experiment, we first analyze the curvature distributions of YaleB database and USPS database which are shown in Figure 1. From Figure 1, we can see that the embedded manifold of USPS database is highly curved in higher dimensional Euclidean space. The curvature value in each data point of USPS database is almost higher than $0.5$. One reason is that the handwritten digits from different classes are vary greatly. However, for Extended YaleFace B database, the curvature distribution of each point is in the range of $0$ to $5\times 10^{-3}$. It means that the local geometric structure of YaleB database is close to flat space.

\begin{sidewaystable}
	\centering  
	\begin{tabular}{|l|c|c|c|c|}  
		\hline
		YFB DB &YFB-trn10/tst40 &YFB-trn20/tst30 &YFB-trn30/tst20 &YFB-trn40/tst10 \\ \hline
		PCA [12] &$44.26 \pm 1.2$ &$52.31 \pm 2.8$ &$55.72 \pm 1.9$ &$62.97 \pm 1.7$ \\ 
		LPP [4] &$29.70 \pm 2.5$ &$53.54 \pm 1.5$ &$61.84 \pm 1.8$ &$70.29 \pm 1.4$ \\ 
		LEP [1] &$34.43 \pm 2.1$ &$53.87 \pm 1.3$ &$61.28 \pm 1.8$ &$68.62 \pm 1.5$  \\ 
		LLE [6] &$43.01 \pm 1.9$ &$51.72 \pm 1.8$ & $50.56 \pm 2.3$ & $60.42 \pm 1.7$  \\ 
		CA-LEP &$45.03 \pm 1.4$ &$55.02\pm 1.2$ &$63.38 \pm 1.6$ &$73.87 \pm 1.3$ \\
		CA-LLE &$43.21 \pm 1.5$ &$52.31 \pm 0.9$ &$51.64 \pm 1.3$ &$61.39 \pm 1.2$\\
		\hhline{=====}
		USPS DB &USPS-trn300/tst400 &USPS-trn400/tst300 &USPS-trn500/tst200 &USPS-trn600/tst100  \\ \hline
		PCA [12] &$86.68 \pm 1.3$ &$84.71 \pm 1.2$ &$86.40 \pm 1.5$ &$87.20 \pm 1.1$ \\ 
		LPP [4] &$86.45 \pm 1.4$ &$88.62 \pm 2.1$ &$89.84 \pm 1.3$ &$90.14 \pm 1.5$ \\ 
		LEP [1] &$91.25 \pm 1.2$ &$91.89 \pm 0.8$ &$92.03 \pm 0.7$ &$92.81 \pm 1.3$  \\ 
		LLE [6] &$89.72 \pm 1.3$ &$90.97 \pm 1.2$ &$91.30 \pm 0.8$ &$92.48 \pm 0.6$ \\
		CA-LEP &$93.31 \pm 1.1$ &$93.52 \pm 1.6$ &$94.14 \pm 1.4$ &$94.52 \pm 1.2$ \\
		CA-LLE &$91.46 \pm 0.9$ &$92.08 \pm 1.3$ &$92.31 \pm 1.7$ &$93.04 \pm 1.5$ \\ \hline
	\end{tabular}	
\caption{Classification performance of YFB DB, USPS DB, together with the comparison results for CAML (CA-LEP and CA-LLE) and traditional manifold learning algorithms PCA, LPP, LEP, and LLE. }
\end{sidewaystable}
	
In the second step of this experiment, we compare our algorithm with traditional manifold learning algorithms under these two databases. The whole experiment design is shown as follows: First, we use manifold learning algorithms to reduce the dimension of databases. Second, in the low dimension space, we use \textsl{Nearest Neighbor Classifier} to test the classification accuracies of these two databases. 
	
For YFB database, we choose $50$ images per subject totally $1900$ images in our experiment. We totally do the experiment four times by each algorithm. In each experiment, we randomly choose $p\left(p=10,20,30,40\right)$ images per subject as the training dataset, the rest $50-p$ images per subject as the testing dataset respectively. The classification accuracy results of different algorithms are shown in Table 3 upper part. The main purpose of this experiment is to find the improvement of our curvature-aware manifold learning algorithm compared with traditional manifold learning algorithms without considering the curvature information. From Table 3, the classification results of manifold learning algorithms mostly outperform the linear dimension reduction algorithm PCA. In addition, the classification results of LPP and LEP are especially higher than LLE. One main reason is that LLE assumes the local patches of data points are linear space and uncovers the linear combination relationship. Among all these classification results, we especially propose to analyze the comparisons between traditional manifold learning algorithms and our curvature-aware manifold learning algorithm. After adding curvature information to LLE, the classification results of CA-LLE slightly outperform LLE. One main reason is that the curvature distribution of YFB database is close to zero. In all, the performance of our curvature-aware manifold learning algorithm is better than all the other algorithms.
	
For USPS Database, we choose $700$ images per subject in this experiment. We also do the experiments four times by each algorithm respectively. As the same method with YFB DB, we randomly choose $p, \left(p = 300, 400, 500, 600 \right)$ respectively image sets per subject for training, the rest for testing. At each time, we choose the different data set as the training set, the rest as the testing set. The whole classification results of different algorithms are shown in Table 3 lower part. From these results, we can see that the classification results of traditional manifold learning algorithms outperform PCA in any case. For our curvature-aware manifold learning, we consider the curvature information of data points. Compared with traditional manifold learning algorithms, the classification accuracies of our algorithm are higher than the other algorithms. Among these results, our focus is to compare the classification accuracies between LEP, LLE and CA-LEP, CA-LLE. It can be observed that our method significantly outperforms them. 
	
Above all, when adding the curvature information of data points into manifold learning, the results of our algorithm outperform other traditional manifold learning algorithms in any case.

\section{Conclusions and Future Works}
	
To precisely describe the continuous change of point cloud, one critical step of manifold learning is to assume the dataset distributed on a lower dimensional embedded manifold. Then they use the mathematical theoretical knowledge of manifold to deal with these datasets, such as dimensionality reduction, classification, clustering, recognition and so on. Whether the manifold structure is uncovered exactly or not directly impacts the learning results. Traditional MAL algorithms just consider the distance metric. However, general Riemannian manifold may be not isometric to Euclidean space. So our method aims to excavate the higher order geometric quantity Riemannian curvature of Riemannian sub-manifold and uses curvature information as well as distance metric to uncover the intrinsic geometric structure of local patches. The extensive experiments have shown that our method is more stable compared with other traditional manifold learning algorithms. It is the first time to try to add curvature information on high dimensional data points for dimensionality reduction. 
	
In the future, this work will try to use Ricci flow to dynamically uncover the intrinsic curvature structure of sub-manifold. Next we will further study the Ricci flow theory, and apply these research results on manifold learning field.
	
\section*{Aknowledgments}
	
This work is supported by the National Key Research and Development Program of China under grant 2016YFB1000902, NSFC project No.61232015, No.61472412, No.61621003, the Beijing Science and Technology Project: Machine Learning based Stomatology and Tsinghua-Tencent-AMSS-Joint Project: WWW Knowledge Structure and its Application.




\end{document}